\documentclass[journal]{IEEEtran}

\usepackage{amssymb}
\usepackage{amsmath}
\usepackage{dcolumn}
\usepackage{graphics}
\usepackage{hyperref}
\usepackage{multirow}
\usepackage{booktabs}
\usepackage{graphicx}
\usepackage{algorithm}
\usepackage{algorithmicx}
\usepackage{algpseudocode}

\begin{document}
		
\title{AIP: Adversarial Iterative Pruning Based on Knowledge Transfer for Convolutional Neural Networks}

\author{Jingfei~Chang,
	Yang~Lu,
	Ping~Xue,
	Yiqun~Xu,
	and~Zhen~Wei
	\thanks{Corresponding author: Yang Lu.}
	\thanks{J. Chang, Y. Lu, P. Xue, Y. Xu and Z. Wei are with the School of Computer Science and Information Engineering, Hefei University of Technology, Hefei 230009, China (e-mail: cjfhfut@mail.hfut.edu.cn).}}

\maketitle
		
\begin{abstract}
With the increase of structure complexity, convolutional neural networks (CNNs) take a fair amount of computation cost. Meanwhile, existing research reveals the salient parameter redundancy in CNNs. The current pruning methods can compress CNNs with little performance drop, but when the pruning ratio increases, the accuracy loss is more serious. Moreover, some iterative pruning methods are difficult to accurately identify and delete unimportant parameters due to the accuracy drop during pruning. We propose a novel adversarial iterative pruning method (AIP) for CNNs based on knowledge transfer. The original network is regarded as the teacher while the compressed network is the student. We apply attention maps and output features to transfer information from the teacher to the student. Then, a shallow fully-connected network is designed as the discriminator to allow the output of two networks to play an adversarial game, thereby it can quickly recover the pruned accuracy among pruning intervals. Finally, an iterative pruning scheme based on the importance of channels is proposed. We conduct extensive experiments on the image classification tasks CIFAR-10, CIFAR-100, and ILSVRC-2012 to verify our pruning method can achieve efficient compression for CNNs even without accuracy loss. On the ILSVRC-2012, when removing 36.78\% parameters and 45.55\% floating-point operations (FLOPs) of ResNet-18, the Top-1 accuracy drop are only 0.66\%. Our method is superior to some state-of-the-art pruning schemes in terms of compressing rate and accuracy. Moreover, we further demonstrate that AIP has good generalization on the object detection task PASCAL VOC.
\end{abstract}

\begin{IEEEkeywords}
Convolutional neural networks, network compression, knowledge transfer, adversarial game, image classification, object detection.
\end{IEEEkeywords}

\IEEEpeerreviewmaketitle

\section{Introduction}

\IEEEPARstart{S}{ince} the emergence of deep neural networks (DNNs) \cite{ISI:000237698100002A01,ISI:000239308600057A02,ISI:000355286600030A03}, due to the less labeled data, poor hardware storage and computing power, it has not been able to completely release the performance. As the number of labeled datasets keeps springing up, as well as the development of high-performance hardware such as GPU and TPU, DNNs have achieved great success in the fields of scientific research and engineering. As the main component, the CNNs achieve excellent performance in extracting image features combined by virtue of the parameter sharing and translation invariance characteristics. At present, CNNs has received extensive attention in computer vision tasks such as image classification \cite{INSPEC:17133663A04}, object detection \cite{ISI:000450913100006A05}, semantic segmentation \cite{ISI:000380414100170A06}, style transfer \cite{ISI:000425498401060A07}, and super-resolution images \cite{ISI:000345525100013A08}, moreover its performance is significantly better than the traditional methods. However, as image and video tasks become more and more complex, the scale and classes of CNNs are gradually increasing. Although this can achieve better accuracy, it also extends the cost of hardware and computing power for network deployment, which limits the application of high-performance CNNs on resource-constrained devices. On the other hand, some works \cite{ISI:000450913101044A09,shih2019realA10,DBLP:journals/ieeejas/SunLSLLL21A11} have shown that existing CNNs have a certain degree of parameter redundancy, which provides background support and a theoretical basis for network compression.

The existing CNNs compressing methods mainly consist of optimizing the calculation methods of convolution \cite{INSPEC:16989755A12} and designing network compression algorithms. In a nutshell, the network compressing is committed to reducing the number of parameters and FLOPs as more as possible in the case of guaranteeing network performance. Mainstream algorithms include network pruning, quantification, low-rank decomposition, and knowledge distillation. Network pruning can delete unimportant neurons or connections in the network. While quantization refers to replacing the full-precision parameters in the original model with 8bit, 4bit, or even binary weights and activations to form a binary network to achieve the high amplitude of compression and acceleration. And low-rank decomposition refers to decomposing the original convolution tensor into low-rank tensors, thereby reducing the number of convolutional operations and accelerating the network. Knowledge distillation utilizes the rich information learned by the large-scale network to guide the training of the small network, which can obtain performance close to the large model. Inspired by the above approaches, we consider designing some strategies in the phase of pruning to efficiently optimize the continuously compressed network. In this way, it can quickly restore the accuracy to perform the iterative pruning in a few training epochs, and it can ensure that the network after pruning has almost no performance loss. The motivation of our method is two-fold. First, \cite{komodakis2017payingA13} demonstrates that feature maps of the large network can pay more accurate attention to the object than the small network through extensive experiments. Secondly, \cite{DBLP:conf/cvpr/LinJYZCYHD19A14} introduces GANs to optimize the network compressing, but it adds a mask for pruning, which increases the cost of the network pruning, and a separate optimization for this parameter is required.

In this paper, we propose an adversarial iterative pruning method based on knowledge transfer. We define the original network as the teacher and the pruned network as the student. First, using the intermediate feature maps of the two networks to construct the attention map to transfer the information learned by the teacher during training to the student. And then the knowledge distillation is introduced so that the output probability of the student will gradually approach that of the teacher. After that, we construct a shallow neural network as a discriminator, making the output features of the two networks conduct an adversarial game. The above strategies act on the student during network pruning. Finally, an iterative pruning strategy based on the importance of channels is designed, which eliminates the unimportant parameters in the network through the threshold. And iterative pruning is carried out during the training phase of the student network. To do so, the pruned network can recover accuracy by training a few epochs after each pruning operation, which can provide more accurate guidance for the judgment of the importance of parameters in the next pruning step, and shorten the whole pruning phase. The final compact network is retrained according to the above optimization method to restore the experimental accuracy.

To demonstrate the effectiveness of our AIP, we prune VGGNet \cite{DBLP:journals/corr/SimonyanZ14aA17}, ResNet \cite{DBLP:conf/cvpr/HeZRS16A18} and GoogLeNet \cite{DBLP:conf/cvpr/SzegedyLJSRAEVR15A19} on the image classification datasets CIFAR-10, CIFAR-100 \cite{krizhevsky2009learningA15} and ILSVRC-2012 \cite{DBLP:journals/ijcv/RussakovskyDSKS15A16}. Moreover, we further perform experiments on the SSD \cite{ISI:000389382700002A21} on the object detection dataset PASCAL VOC \cite{ISI:000348345400006A20}. The results manifest that without harming overall performance it is possible to compress and accelerate the CNNs using the proposed pruning method in this paper. On the CIFAR-10, when removing 97.22\% of the parameters and 96.57\% of the FLOPs of the VGG-16, the classification accuracy can still reach 90.29\%. In addition, when the compression rate of SSD exceeds 50\%, the performance loss is still less than 1.00\%.

The proposed AIP can be applied to many convolutional networks in image classification tasks, and it also shows good generalization in object detection. The existing network compressing method can combine with our knowledge transfer and adversarial strategy to increase the accuracy of the compressed network. What's more, because no sparseness was introduced, AIP does not require the assistance of additional sparse matrix operations and acceleration libraries. And the entire pruning process can be achieved only by controlling one parameter, which notably reduces labor intervention and can perform automatic compression and acceleration. If adopting the larger CNN as the teacher network, the efficiency of pruning and the performance of the compressed network can be further improved.

The main contributions of our work are as follows:
\begin{itemize}
	\item This paper proposes a guidance strategy for network pruning based on knowledge transfer. We use attention transfer and knowledge distillation to accelerate and correct the learning during pruning so that the network can accurately identify unimportant parameters to eliminate.
	
	\item We design an adversarial iterative pruning method. The output features of the original network and the pruned network play games via a discriminator. At the same time, iterative pruning is performed during training to enhance the accuracy recovery speed of the compact network.
	
	\item This paper presents a channel pruning method based on the importance of feature maps. During the training phase, the channel importance score is constructed according to the intermediate feature maps, and then the unimportant parameters and connections smaller than a preset threshold are deleted.
	
	\item  We demonstrate the effectiveness of the method on CIFAR-10, CIFAR-100, ILSVRC-2012 with extensive experiments. Moreover, the results on the object detection dataset PASCAL VOC further verifies our AIP has superior generalization in CNNs compressing and accelerating. The ablation analysis manifests that the adjustment of hyperparameter can stably control the pruning rate.
	
\end{itemize}

\begin{figure*}[h]
	
	\centering
	
	\includegraphics[width=16cm]{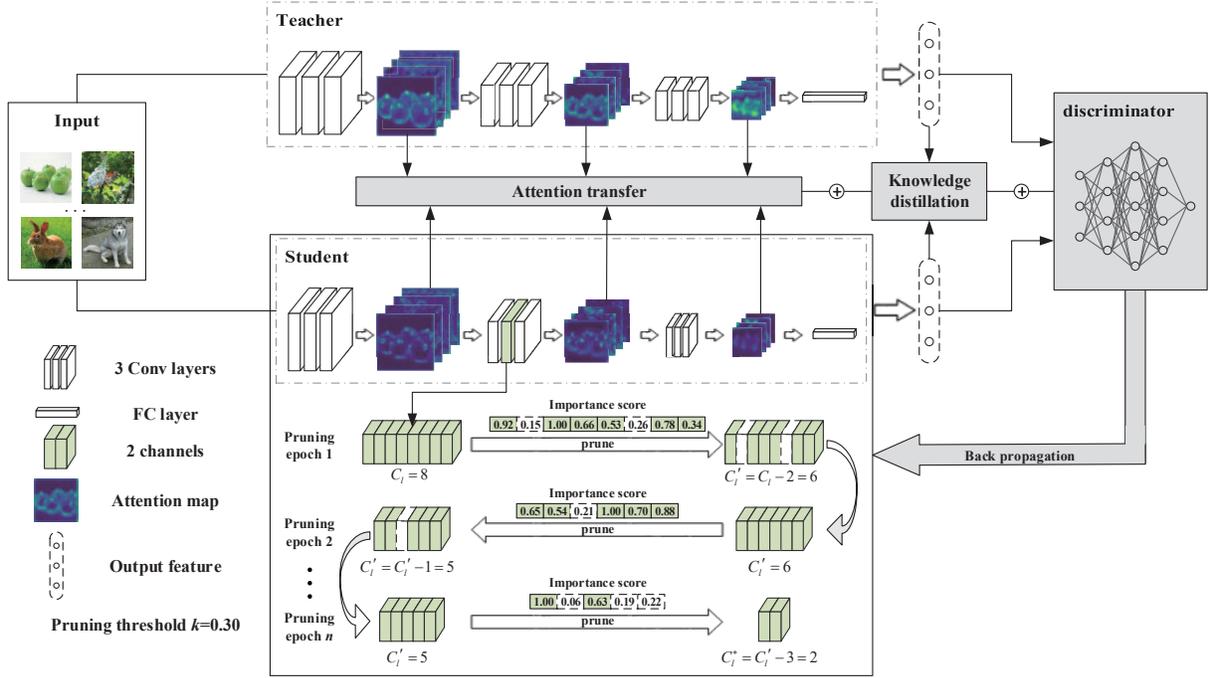} 
	
	\caption{The adversarial iterative pruning (AIP) framework. (This figure is best viewed in color and zoomed in.)}
	\label{img1}
	
\end{figure*}

\section{Related Works}
At present, convolutional network compression has received widespread attention from both academia and industry. Many methods with significant effects such as pruning, quantification, low-rank decomposition, and knowledge distillation have emerged. The channel pruning scheme AIP proposed in this paper draws lessons from the ideas of network pruning, knowledge distillation, and generative adversarial network. The related works are presented as follows:

\subsection{Network Pruning} Network pruning is to remove the relatively redundant weights or filters according to the importance of parameters in the CNN to compress and accelerate the network based on ensuring the accuracy of the task. The key of pruning is to determine the evaluation criteria of the importance of parameters and then design an effective pruning strategy. Some existing methods are based on the magnitude of parameters, such as \cite{ISI:000450913101044A09} using the value of weight to measure the redundancy of connections. \cite{DBLP:conf/iclr/0022KDSG17A22} deletes the filters with the smaller L1 norm. \cite{DBLP:conf/ijcai/HeKDFY18A23} prunes networks according to the L2 norm of filters. \cite{ISI:000425498402086A24} uses the scaling factor of the batch normalization layers as the evaluation standard of parameter importance. \cite{DBLP:conf/cvpr/LinJWZZ0020A25} utilizes the rank of the feature map matrix to judge how much information it contains. \cite{DBLP:conf/ijcai/LiMXL20A26} slims CNN through the diversity and similarity of feature maps. \cite{tang2021manifoldA27} fits the input complexity and feature similarity to the pruned network space to dynamically discard redundant filters. \cite{wu2021pruningA28} uses the product of filter sparsity and feature dispersion to measure their importance. Some pruning strategies are based on the impact of deleted parameters on performance drop. For example, \cite{DBLP:conf/cvpr/Yu00LMHGLD18A29} measures the importance of pruning neurons by minimizing the reconstruction error of the second-to-last layer in front of the final classification layer. \cite{DBLP:conf/iclr/LeeAT19A30} introduces connection sensitivity to evaluate the importance of structure, and the pruning is implemented in the parameter initialization stage before training. \cite{DBLP:conf/cvpr/GuoWLY20A31} samples the channel pruning as a Markov process that is optimized using standard regularization loss and model parameters or FLOPs budget regularization. \cite{DBLP:conf/nips/YouYYM019A32} multiplies the intermediate feature map with a scale factor, and then estimates the accuracy loss caused by the scale factor set to zero to determine the importance of the relevant filters. \cite{guo2020channelA33} reconstructs the cropped feature and observe its impact on the classification loss to carry out layer-by-layer channel pruning. Other pruning approaches combine existing advanced algorithms to compress the network, such as \cite{DBLP:conf/eccv/HeLLWLH18A34} using reinforcement learning to search for better pruning strategies. \cite{liu2019metapruningA35} combines meta-learning to find compact networks with better performance. \cite{DBLP:conf/ijcai/LinJZZW020A36} formulates the search of optimal pruned structure as an optimization problem and integrate the ABC algorithm to solve it in an automatic manner. \cite{wang2020networkA37} proposes a channel pruning scheme based on sparse learning and genetic algorithm. \cite{ding2020pruneA38} generates a global network pruning strategy using long short-term memory. To simplify the pruning process as much as possible, we design an iterative pruning strategy based on the importance of feature maps with almost no decrease in accuracy.

\subsection{Knowledge Transfer} Knowledge transfer utilizes a pre-trained high-performance teacher network to guide a smaller student network, thereby improving the experimental accuracy of the small network. \cite{DBLP:conf/nips/BaC14A39} uses the input of the final softmax layer to represent the knowledge learned by the teacher network to supervise the training of the student network. \cite{INSPEC:16231013A40} introduces temperature T to the output of the softmax layer and then trains together as a soft label with the real target. The FitNet proposed in \cite{DBLP:journals/corr/RomeroBKCGB14A41} applies not only the output of the teacher network but also its intermediate feature to jointly optimize the training process of the student network. This can train a deep and narrow student network while enhancing its generalization ability. \cite{komodakis2017payingA13} proposes attention transfer that using the attention maps in the teacher network to deliver the information of the teacher network’s attention to the student network and improve the performance. \cite{yim2017giftA42} considers the inner product matrix between the middle layers as the domain knowledge to guide the student network and apply it to different tasks. We mainly introduce the information representation of feature maps of the teacher network to direct the student network to achieve precise pruning.

\subsection{Generative Adversarial Network}

\cite{ISI:000452647101094A43} proposes the generative adversarial networks (GANs) inspired by the idea of the zero-sum game. GANs are mainly composed of a generator and a discriminator. The purpose of the generator is to generate vivid samples to deceive the discriminator, and the discriminator should try its best to distinguish between the real and fake samples. Both of them continuously improve their abilities through confrontation training until they reach the state of Nash equilibrium. \cite{DBLP:journals/corr/RadfordMC15A44} introduces DCGANs, which use deep CNNs instead of the multilayer perceptron in the original GANs to broaden the application in the image field. To solve the problem of unstable training and uncontrollable output of GANs, \cite{mirza2014conditionalA45} proposes Conditional Generative Adversarial Networks (CGANs), and \cite{odena2017conditionalA46} proposes AC-GANs, which use labeled images classification tasks to improve the generalization of models. In addition, \cite{DBLP:conf/nips/DentonCSF15A47,zhao2016energyA48,shaham2019singanA49} and some other works have contributed to improving the performance and stability of GANs. Our AIP makes the outputs of the student and teacher network game with each other and then feeds back to the student network through backpropagation to optimize the pruning strategy.


\section{Proposed Method}

In this paper, we propose an adversarial iterative network pruning method based on knowledge transfer. The original unpruned network is regarded as a teacher network whose middle attention maps and the output features are used to guide the iterative pruning of the student network. At the same time, we apply the game theory to introduce the discriminator for further improving the learning ability of the student network during compressing. In this scenario, the unimportant parameters are exactly deleted at each pruning step, and then the performance of the compact network is restored through retraining. In this section, we firstly present the overall framework of our method. Then the adversarial strategy based on knowledge transfer and the iterative channel pruning scheme are introduced respectively. Finally, the implementation details for pruning different convolutional neural networks are explained.

\subsection{The Adversarial Iterative Pruning Framework}

Fig.\ref{img1} shows the overall framework of our pruning method. Firstly, the labeled training images are input into the teacher network and the student network respectively. Each image will produce intermediate feature maps and the final output feature in two networks. We select three-pair feature maps of the two networks to transfer attention and perform knowledge distillation. It can be seen from the figure that for the same input sample, the attention maps generated by the pruned student network have obviously weaker interest in the classification object than the original teacher network. Afterwards, a shallow neural network is used as a discriminator to make the two outputs play a game, which further improves the accuracy of the pruned network. Finally, combining the information generated by the three parts aforementioned to conduct back-propagation on the student network for training. Meanwhile, we perform iterative pruning to compress and accelerate the original convolutional network.

\subsection{Adversarial Scheme Based on Knowledge Transfer}

Given a convolutional neural network with $L$ layers, we refer to $C=\left( {{C}_{1}},{{C}_{2}},\ldots ,{{C}_{L}} \right)$ as the original network structure, where ${C}_{L}$  is the number of channels in the $l$-th layer. $W\in {{\mathbb{R}}^{{{C}_{out}}\times {{C}_{in}}\times k\times k}}$ is the weight of the filter, where ${C}_{out}$ is the number of output channels, ${C}_{in}$ is the number of input channels, and $k\times k$ is the size of the filter. The feature map generated in the $l$-th layer is $M\in {{\mathbb{R}}^{{{C}_{l}}\times w\times h}}$, where $w$ and $h$ are width and height of the feature map. With the input sample $x$, the output produced by the original network is defined as ${{f}_{T}}\left( x,{{W}_{T}} \right)$, and the output of the pruned network is defined as ${{f}_{S}}\left( x,{{W}_{S}} \right)$. $D\left( f\left( x,W \right),{{W}_{D}} \right)$ is the discriminator, where $f\left( x,W \right)$ is the output of the teacher or the student network.

\subsubsection{Knowledge Transfer}

The intermediate feature map of CNNs is the concrete or abstract representation extracted by the filters from the input images, which shows the objects that the network pays attention to when treating specific tasks. For image classification, the feature maps will highlight the target to be classified and weaken the background and irrelevant objects to obtain a more reliable classification result. Therefore, whether the feature map can precisely pay attention to the goal and how strong the attention is are especially important to the performance of the network. It can be seen from Fig.\ref{img1} that the feature maps of the student network pay less attention to the target, which will seriously affect the correctness of pruning and the accuracy of the compressed network. Because of this, we introduce knowledge transfer in the pruning process, including attention transfer and knowledge distillation.

We select three layers with the different dimensions of feature maps and integrate the feature maps in the same layer to form an attention map to perform attention transfer. Specifically, for the ${{C}_{l}}$ intermediate feature maps of the $l$-th layer, the attention map is constructed using the Eq.\ref{eq1}:
\begin{equation}
{{A}^{l}}\left( M \right)=\sum\limits_{i=1}^{{{C}_{l}}}{|{{M}_{i}}{{|}^{2}}}.
\label{eq1}
\end{equation}

The attention map produced in this way can get the attention area of the input sample, and on the other hand, it can also represent the amount of information learned in the layer. Then as shown in Eq.\ref{eq2}, after regularizing the attention map of two networks, we use the L2 norm of their difference to construct the attention transfer loss ${{\mathcal{L}}_{AT}}$ of the student network.
\begin{equation}
{{\mathcal{L}}_{AT}}=\sum\limits_{l\in \mathsf{\mathcal{H}}}{{{\left\| \frac{A_{S}^{l}}{||A_{S}^{l}|{{|}_{2}}}-\frac{A_{T}^{l}}{||A_{T}^{l}|{{|}_{2}}} \right\|}_{2}}}
\label{eq2}
\end{equation}

\begin{figure*}[htbp]
	
	\centering
	
	\includegraphics[width=15cm]{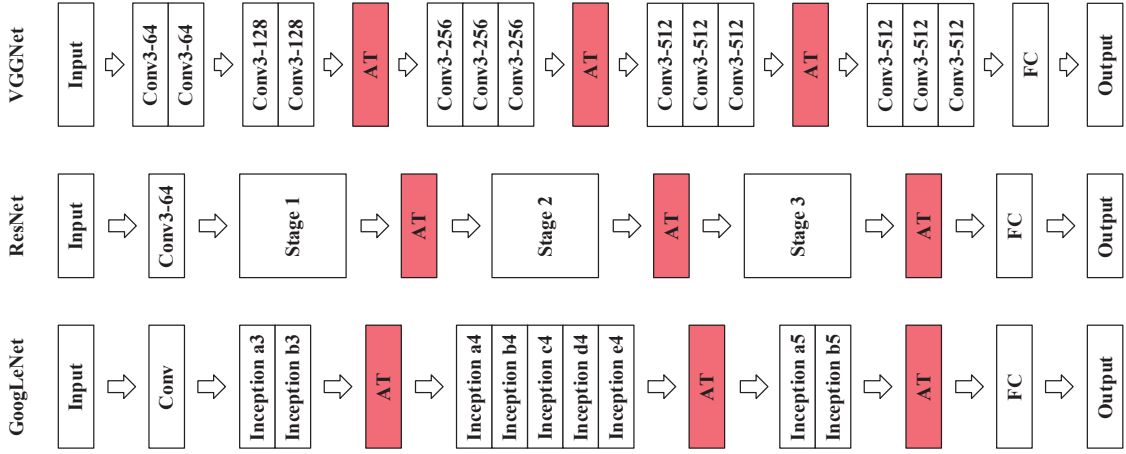} 
	
	\caption{The positions of VGGNet, ResNet, and GoogLeNet for attention transfer.}
	\label{img2}
	
\end{figure*}

Where $\mathcal{H}$ indicates all the teacher-student convolutional layer pairs that perform attention transfer, $A_{S}^{l}$ is the attention map in the $l$-th layer of the student network, and $A_{T}^{l}$ is the corresponding attention map of the teacher network. The interest of the student can be made as close as possible to that of the teacher in the inference process through the attention transfer loss. In the experiments, we prune VGGNet, ResNet, and GoogLeNet. The specific implementation positions of attention transfer in these three networks are plotted in Fig.\ref{img2}, where the highlighted AT refers to the attention transfer operation.

When dealing with image classification tasks, the output of the convolutional network is the probabilities of each category, so we apply the output features of the teacher to guide the student. In this way, we can perform more accurate pruning and improve the performance of the student network at the same time. Regarding the outputs of the teacher and the student network ${{f}_{T}}\left( x \right)$ and ${{f}_{S}}\left( x \right)$, this paper introduces temperature ${{T}_{emp}}$ which draws on the idea of \cite{INSPEC:16231013A40} to smooth the two outputs as shown in Eq.\ref{eq3} and Eq.\ref{eq4}. Hence the classification probability of the student for each category can be as similar to the teacher network as possible to avoid the probability of all incorrect classification tends to zero, and result in a smoother category probability distribution.
\begin{equation}
p\left( x \right)={{F}_{soft\max }}\left( {{{f}_{S}}\left( x \right)}/{{{T}_{emp}}}\; \right)
\label{eq3}
\end{equation}
\begin{equation}
q\left( x \right)={{F}_{soft\max }}\left( {{{f}_{T}}\left( x \right)}/{{{T}_{emp}}}\; \right)
\label{eq4}
\end{equation}
Where ${{F}_{soft\max }}\left( \centerdot  \right)$ refers to softmax function. Then the KL divergence of $p\left( x \right)$ and $q\left( x \right)$ is calculated according to Eq.\ref{eq5}, and the accuracy of the student can be improved by reducing the divergence during training. 
\begin{equation}
{{D}_{KL}}\left( p\parallel q \right)=\sum\limits_{i=1}^{n}{\left[ p\left( x \right)log\left( p\left( x \right) \right)-p\left( x \right)log\left( q\left( x \right) \right) \right]}
\label{eq5}
\end{equation}

At the same time, to better correct the output of the student network, the cross-entropy loss ${{L}_{CE}}\left( {{W}_{S}} \right)$ between the output features of the student and the real labels is added to the above divergence, which is regarded as the knowledge distillation loss ${{L}_{KD}}\left( {{W}_{S}} \right)$.
\begin{equation}
{{\mathcal{L}}_{KD}}\left( {{W}_{S}} \right)=\alpha \cdot {{\mathcal{L}}_{KL}}\left( {{W}_{S}} \right)+(1-\alpha ){{\mathcal{L}}_{CE}}\left( {{W}_{S}} \right)
\label{eq6}
\end{equation}
Where, $\alpha$ is the weight between the two losses of KL divergence and cross entropy. And ${{L}_{KL}}\left( {{W}_{S}} \right)$ is formulated in Eq.\ref{eq7}. In order to make the effect of these two losses roughly under the same magnitude, we multiply ${{D}_{KL}}$ by $T_{emp}^{2}$.
\begin{equation}
{{\mathcal{L}}_{KL}}\left( {{W}_{S}} \right)\text{=}T_{emp}^{2}\centerdot {{D}_{KL}}\left( p\parallel q \right)
\label{eq7}
\end{equation}

Through the above two methods, the intermediate and the final classification information learned by the original network can be thoroughly transmitted to the pruned student network. In this scenario, we can not only ensure that the student accurately finds the unimportant parameters for the corresponding task but also restore the network's performance through a few training epochs after each pruning step to achieve iterative pruning during the training process. 

\subsubsection{Adversarial Game}

The above-mentioned knowledge transfer has achieved the effective delivery of semantic information from the teacher to the student network. On this basis, we found that introducing an adversarial game strategy can further improve the final output performance of the student and the recovery speed of accuracy in the iterative pruning. Hence, this paper constructs a shallow neural network as the discriminator and makes the outputs of the student and the teacher network play an adversarial game on it. In this way, the output of the student network may closer approach that of the teacher network, then the response of the discriminator to the output features of the student will increase. Therefore, the adversarial loss of the student network is defined as follows:
\begin{equation}
{{L}_{A}}({{W}_{S}})={{E}_{{{f}_{S}}(x)\sim {{p}_{S}}(x)}}\left[ log\left( 1-D\left( {{f}_{S}}\left( x,{{W}_{S}} \right),{{W}_{D}} \right) \right) \right],
\label{eq8}
\end{equation}
where, ${{p}_{S}}(x)$ represents the feature distribution of the student network. Combined with the knowledge transfer loss in the previous section, the training loss of the student network in the proposed pruning method consists of the following three parts:
\begin{equation}
{{\mathcal{L}}_{S}}\left( {{W}_{S}} \right)={{\mathcal{L}}_{A}}\left( {{W}_{S}} \right)+{{\mathcal{L}}_{AT}}\left( {{W}_{S}} \right)+{{\mathcal{L}}_{KD}}\left( {{W}_{S}} \right).
\label{eq9}
\end{equation}

The discriminator needs to be continuously trained to distinguish whether the input is from the teacher or the pruned network. For the output features from the teacher network, the discriminator should produce a positive response, while for the output features generated by the student network, the discriminator should treat it as the pseudo sample. To be specified, the loss of the discriminator during training is defined as follows:
\begin{equation}
{\begin{array}{l}
	{{\mathcal{L}}_{D}}({{W}_{D}})={{E}_{{{f}_{T}}(x)\sim {{p}_{T}}(x)}}\left[ log\left( 1-D\left( {{f}_{T}}\left( x,{{W}_{T}} \right),{{W}_{D}} \right) \right) \right]\\
	+{{E}_{{{f}_{S}}(x)\sim {{p}_{S}}(x)}}\left[ log\left( D\left( {{f}_{S}}\left( x,{{W}_{S}} \right),{{W}_{D}} \right) \right) \right]
\end{array}
\label{eq10}
}
\end{equation}
where, ${{p}_{T}}(x)$ represents the feature distribution of the teacher network.

The discriminator and the pruned network are alternately optimized in each training epoch to accelerate the performance improvement of the student network. In addition, we integrate the attention transfer and knowledge distillation, so that the accuracy of the compact network can be regained only after a few training epochs and then the next pruning will be conducted. Accordingly, the entire network pruning process becomes more compact and accurate. Moreover, our method can significantly improve the accuracy of the network after pruning. Extensive experiments have shown that even in the case of a considerable compressing rate, the performance of the pruned network after retraining can still reach or exceed that of the original network.

\begin{figure*}[htbp]
	
	\centering
	
	\includegraphics[width=18cm]{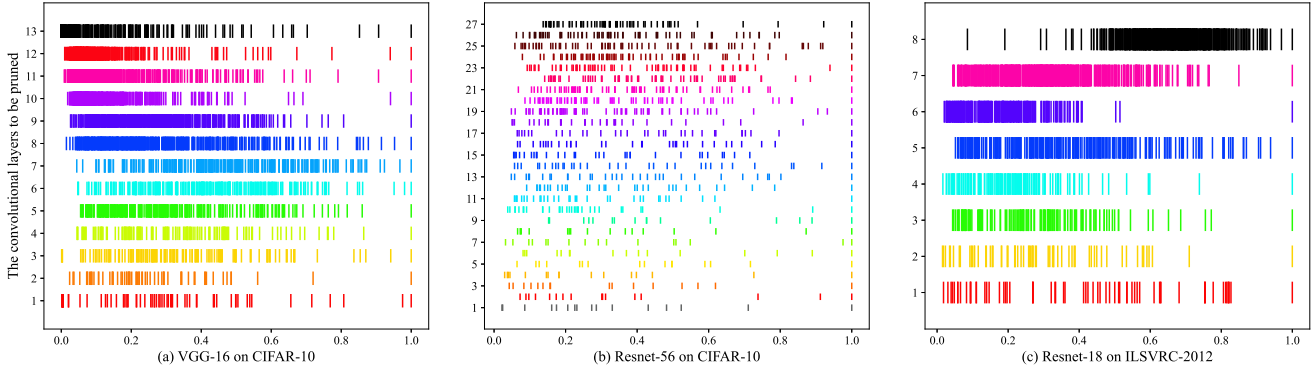} 
	
	\caption{The importance score distribution density of the feature map in the pruned layers.}
	\label{img3}
	
\end{figure*}

\subsection{Iterative Pruning}

According to the optimization method proposed above, we introduce the iterative pruning strategy of the convolutional neural network in this section. To reduce the complexity of the network pruning, an effective pruning method based on the magnitude of parameters should be designed. Considering the smaller parameters contribute less to the backpropagation, wherefore its influence on the accuracy of the network may be limited and can be discarded to reduce parameter redundancy. Most of the existing methods delete unimportant parameters directly based on the L1, L2 norm, or other magnitude of the filters and feature maps. However, this depends on a relatively uniform distribution of feature map’s magnitude. Otherwise, when the pruning threshold is unreasonable, it will cause enormous differences in the pruning rate of layers, which will seriously affect the final performance of the network. Here, we analyze the L1 norm for the feature maps of VGG-16 and Resnet-56 on the CIFAR10 and Resnet-18 on the ILSVRC-2012. Specifically, we first calculate all the L1 norm of the feature maps in the layers to be pruned and then perform maximum regularization on the feature maps in each layer using Eq.\ref{eq11} to obtain the importance score $m_{l}^{c}$ of each feature map.
\begin{equation}
m_{l}^{c}={{{\left| M_{l}^{c} \right|}_{1}}}/{max\{{{\left| M_{l}^{c} \right|}_{1}}\}}
\label{eq11}
\end{equation}
Where, $c\in \left\{ \left. 1,2,\cdots ,{{C}_{l}} \right\} \right.$ is the index of the every feature map in the $l$-th layer. ${{\left| \centerdot  \right|}_{1}}$ refers to the L1 norm. We visualize the results obtained as Fig.\ref{img3}.

It can be seen from the figure that for CIFAR-10, the importance scores of VGG-16 are generally concentrated between 0 and 0.5. While the importance distribution of Resnet-56 is relatively uniform, but the importance scores in the first few layers are almost between 0 and 0.5. On ILSVRC-2012, the importance of the features for Resnet-18 at each layer is significantly different. The importance scores of the sixth layer are almost between 0 and 0.4, while those of the eighth layer mainly vary from 0.4 to 1. Therefore, directly setting the threshold based on the L1 norm can not achieve ideal compression for all layers in the network. We set pruning factor $k$ to perform on the mean value of the importance scores of the feature maps to determine the final pruning threshold $m_{l}^{p}$ of the $l$-th layer.
\begin{equation}
m_{l}^{p}=k\centerdot \frac{1}{{{C}_{l}}}\sum\limits_{c=1}^{{{C}_{l}}}{m_{l}^{c}}
\label{eq12}
\end{equation}
Where, $k\in \left( 0,1 \right)$ is the parameter used to control the network pruning rate, and it is also the only variable parameter in our proposed method. Redundant parameters in the CNNs can be deleted at each pruning step via the above pruning strategy, and the problem of unbalanced compressing among layers will not occur. The subsequent experiments in this paper also entirely demonstrate the effectiveness and accuracy of our AIP in network compressing and accelerating. Algorithm \ref{alg1} shows the pseudocode of the adversarial iterative pruning method. Given a pre-trained original convolutional network, a compact model $S_{pruned}^{*}$ can be obtained after pruning with the AIP scheme. Finally, we retrain the pruned model from scratch to restore the accuracy of the experiment.

\begin{algorithm}[htbp]
	\caption{Adversarial iterative pruning (AIP)}
	\label{alg:algorithm}
	\textbf{Input}: Training set ${{X}_{train}}=\{{{x}_{1}},{{x}_{2}},\cdots ,{{x}_{n}}\}$ with $n$ samples, teacher model ${{T}_{model}}$, student model ${{S}_{model}}=\{{{C}_{1}},{{C}_{2}},\cdots ,{{C}_{L}}\}$ with weight ${{W}_{S}}$, discriminator ${{D}_{model}}$ with weight ${{W}_{D}}$, learning rate $\eta$, num epochs $N$, epochs of pruning intervals ${{s}_{p}}$, pruning threshold factor $k$\\
	\textbf{Output}: Pruned compact structure ${{S}_{pruned}^{*}}=\{{{C}_{1}^{*}},{{C}_{2}^{*}},\cdots ,{{C}_{L}^{*}}\}$ with wight ${{W}_{S}^{*}}$
	\begin{algorithmic}[1] 
		\State Initialize ${{S}_{model}}$ with weight ${{W}_{S}}$;
		\For{$epoch=1$ to $N$}  
		\State \textbf{$<$ Fix ${S}_{model}$ and update ${D}_{model}$ $>$}
		\State Sample output ${{f}_{S}}\left( x \right)$ of ${{S}_{model}}$;
		\State Sample output ${{f}_{T}}\left( x \right)$ of ${{T}_{model}}$;
		\State Update weight ${{W}_{D}}$ of discriminator ${{D}_{model}}$ via Eq10;
		\State \textbf{$<$ Fix ${{D}_{model}}$ and update ${{S}_{model}}$ $>$}
		\State Update weight ${{W}_{S}}$ of student model ${{S}_{model}}$ via Eq9;
		\State \textbf{$<$ Prune the ${{S}_{model}}$ $>$}
		\If {$epoch\%{s}_{p}==0$}
		\For{$l=1$ to $L$} 
		\State $j=0$;
		\For{$c=1$ to ${C}_{l}$}
		\State Calculate the important score ${m}_{l}^{c}$ via Eq11;
		\State Calculate the pruning threshold ${m}_{l}^{p}$ of the $l$-th layer via Eq12; 
		\If {${m}_{l}^{c}<{m}_{l}^{p}$}
		\State Prune the $c$-th channel and the filters related;
		\State $j=j+1$;
		\EndIf
		\EndFor
		\State ${{C}_{i}^{'}}={C}_{i}-j$;	
		\EndFor
		\State Pruned student structure ${{S}_{pruned}^{'}}=\{{{C}_{1}^{'}},{{C}_{2}^{'}},\cdots ,{{C}_{L}^{'}}\}$;
		\EndIf
		\State ${{S}_{model}}={{S}_{pruned}^{'}}$;
		\EndFor 
	\end{algorithmic}
	${S}_{pruned}^{*}={S}_{model}$.
	\label{alg1}
\end{algorithm}

For most of the existing network pruning methods, the compressed network inherits the weights and bias from the original network to restore the performance as much as possible through fine-tuning. However, when the network pruning rate is remarkable, the accuracy recovery after fine-tuning is not obvious, and the actual performance of the compact network cannot be greatly manifested. \cite{DBLP:conf/iclr/LiuSZHD19A50} makes a surprising observation in structured network pruning that fine-tuning a pruned model only gives comparable or worse performance than training that model with randomly initialized weights. And the experiment results reveal that the pruned architecture itself, rather than a set of inherited important weights, is more crucial to the efficiency in the final model. Our results of pruning VGG-16 on the CIFAR-10 further verify the observation in \cite{DBLP:conf/iclr/LiuSZHD19A50}. In order to fully demonstrate the performance of the compact network, we retrain the pruned network from scratch via the method in section 3.2.1 in our experiments. Specifically, we keep the number of FLOPs consistent before and after pruning. The number of training epochs of the original network is multiplied by the accelerating rate of FLOPs as the retraining epochs of the compressed network. Finally, we compare the accuracy of the pruned network with the original network to draw a conclusion.

\begin{figure}[t]
	
	\centering
	
	\includegraphics[width=8cm]{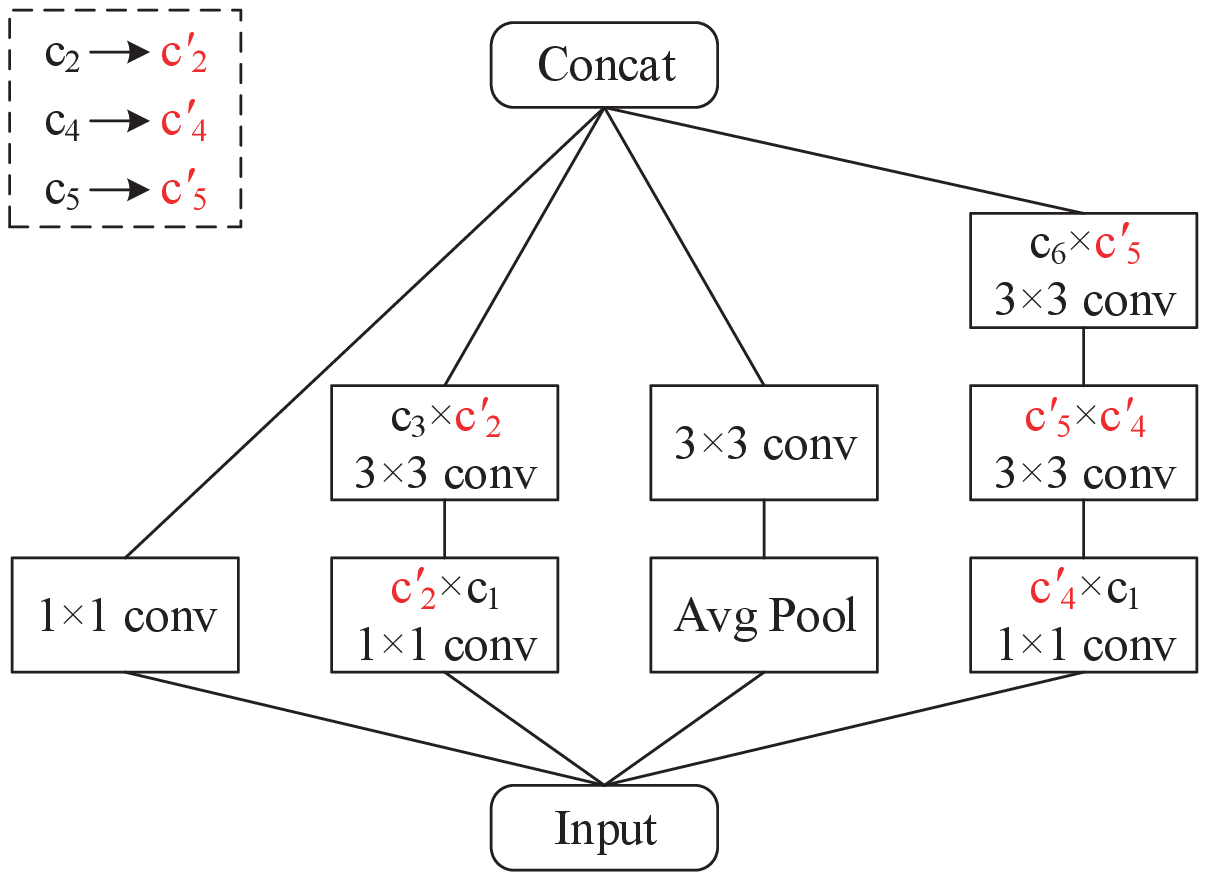} 
	
	\caption{Illustration of pruning Inception V3 module. The black font indicates the number of original channels, and the red font indicates that after pruning.}
	\label{img4}
	
\end{figure}

\subsection{Pruning Strategy for Different CNNs}

Since different CNNs have different network structures, the specific pruning implementation details should also change accordingly. We perform experiments on VGGNet, ResNet, and GoogLeNet. Among them, VGGNet is a common layer-by-layer convolutional network and does not include unusual architecture. Therefore, all layers can be directly pruned without affecting the integrity of the final network structure. ResNet contains customized residual modules, so arbitrarily compressing each layer will destroy the dimension matching of the channels. The basic residual block is composed of two convolutional layers. We only discard the output channels in the first layer, and the input channels in the second layer will also change accordingly. By doing so, the overall dimension of the ResNet is still matched and can be trained correctly after pruning. GoogLeNet is a more complex convolutional network with multiple Inception V3 modules, each of which contains four branches. We cut the branches containing two and three convolutional layers to conduct the compressing and accelerating. The specific structure and pruning scheme of the Inception V3 module is plotted in Fig.\ref{img4}.

\section{Experiments}

We demonstrate the effectiveness of the proposed method by pruning VGGNet, ResNet, and GoogLeNet on the CIFAR-10, CIFAR-100, and ILSVRC-2012. Moreover, we compress SSD via AIP on the PASCAL VOC to analyze its generalization on object detection. All experiments are implemented with Pytorch on NVIDIA TITAN X GPUs. For fairly comparing with the existing pruning methods, the network pre-training and parameter settings use the method presented in \cite{DBLP:conf/cvpr/HeZRS16A18}. Specifically, the pre-training epochs of CNNs on the CIFAR are 160, while on the ILSVRC-2012 are 90. The learning rate is initially set to 0.1 and then decreased by a factor of 10 on half and three-quarter epochs. Stochastic gradient descent (SGD) with momentum is used for backpropagation, and the momentum is 0.9 with a weight decay of 1e-4. In the retraining stage, we adjust the learning rate with the cosine annealing adjustment strategy. The parameter settings during the iterative pruning are as follows. The weight in knowledge distillation is $\alpha$=0.3. On the CIFAR, the total epochs of training are $N$=30, and the pruning interval period is ${{s}_{p}}=10$. On the ILSVRC-2012, the training epochs for pruning are $N$=20, and the pruning interval is ${{s}_{p}}=10$. The pruning threshold factor $k$ is the only parameter that is changed for pruning. In addition, we draw on the neural network composed of three fully-connected layers with the neurons of 128-256-128 in \cite{DBLP:conf/cvpr/LinJYZCYHD19A14} as the discriminator in the adversarial game.

In this section, we compare the proposed method with the existing pruning schemes, among which Li et al. \cite{DBLP:conf/iclr/0022KDSG17A22}, SFP \cite{DBLP:conf/ijcai/HeKDFY18A23}, DCP \cite{DBLP:conf/nips/ZhuangTZLGWHZ18A51}, FPGM \cite{he2019filterA52}, EDP \cite{9246734A53}, CNN-FCF \cite{li2019compressingA54}, CCP \cite{peng2019collaborativeA55}, Taylor-FO-BN \cite{molchanov2019importanceA56}, HRank \cite{DBLP:conf/cvpr/LinJWZZ0020A25}, ManiDP \cite{tang2021manifoldA27} are the state-of-the-art methods. Due to the difference in experimental equipment and environment, the results obtained by different papers also have several differences. In order to make a fair comparison as much as possible, we also mainly compare the decrease of accuracy after pruning according to current methods. The results of these competing methods are reported according to the original article.

\begin{table*}[htbp]	
	\centering
	\caption{Performance comparison of VGG-16 on CIFAR-10. $``$Acc.drop$"$ is the accuracy drop of the pruned network, so a negative number means the compressed model has better performance than the baseline. A smaller number of $``$Acc.drop$"$ is better. $``$Parameters.drop$"$ and $``$FLOPs.drop$"$ are the pruned percentage of the parameters and FLOPs, respectively.}
	\renewcommand{\arraystretch}{1.1}
	{
		\begin{tabular}{c c c c c c c c}			
			\hline 
			Method      & Base Acc/\% & Pruned Acc/\% & Acc.drop/\%  & Parameters/M & Parameters.drop/\% & FLOPs/M & FLOPs.drop/\% \\ 
			\hline 
			\textbf{Baseline}         & 93.60  & $-$    & $-$    & 14.73        & $-$                & 314.59  & $-$           \\ 	
			\textbf{AIP($\textit{k}$=0.3)}  & \textbf{93.60}  & \textbf{94.14}  & \textbf{-0.54}  & \textbf{7.68} & \textbf{47.86}  & \textbf{174.94}  & \textbf{44.39}    \\ 
			Li et al. \cite{DBLP:conf/iclr/0022KDSG17A22}                       & 93.25  & 93.40   & -0.15     & 5.40         & 64.00            & 206.00  & 34.20         \\ 
			Liu et al. \cite{DBLP:conf/iclr/LiuSZHD19A50}                    & 93.63  & 93.78   & -0.15     & 5.40         & 64.00            & 206.00  & 34.20         \\ 
			Zhao et al. \cite{zhao2019variationalA57}                  & 93.25  & 93.18   & 0.07      & 3.92         & 73.34            & 190.00   & 39.10         \\ 	
			ABCPruner \cite{DBLP:conf/ijcai/LinJZZW020A36}                 & 93.02  & 93.08   & -0.06     & 1.67         & 88.68           & 82.81     & 73.68        \\ 
			\textbf{AIP($\textit{k}$=0.5)}  & \textbf{93.60}  & \textbf{93.77}   & \textbf{-0.17}  & \textbf{2.50}  & \textbf{83.03}  & \textbf{60.53}  & \textbf{80.76}          \\ 
			\textbf{AIP($\textit{k}$=0.7)}  & \textbf{93.60}  & \textbf{90.29}   & \textbf{3.31}  & \textbf{0.41}  & \textbf{97.22}  & \textbf{10.79}  & \textbf{96.57}          \\ 	
			\hline 			
	\end{tabular}}	
	\label{tab1}	
\end{table*}

\begin{table*}[t]
	\centering
	\caption{Performance comparison of ResNet-56 on CIFAR-10. The $``$-$"$ indicates that the results are not listed in the original article.}
	\renewcommand{\arraystretch}{1.1}
	{
		\begin{tabular}{c c c c c c}			
			\hline 
			Method  &Baseline Acc/\% & Pruned Acc/\% & Acc.drop/\% & Parameters.drop/\% & FLOPs.drop/\% \\ 
			\hline
			SFP \cite{DBLP:conf/ijcai/HeKDFY18A23}         & 93.59  & 93.89  & -0.30    & $-$    & 14.70   \\
			Li et al. \cite{DBLP:conf/iclr/0022KDSG17A22}   & 93.04  & 93.06  & -0.02    & 13.70    & 27.60   \\
			Liu et al. \cite{DBLP:conf/iclr/LiuSZHD19A50}   & 93.14  & 93.05  &  0.09    & 13.70   & 27.60   \\
			HRank \cite{DBLP:conf/cvpr/LinJWZZ0020A25}       & 93.26  & 93.52  & -0.26  & 29.30   & 16.80   \\
			\textbf{AIP($\textit{k}$=0.4)}  & \textbf{93.18}  & \textbf{94.00}  & \textbf{-0.82}  & \textbf{37.65} & \textbf{46.36}   \\
			SFP \cite{DBLP:conf/ijcai/HeKDFY18A23}         & 93.59  & 93.78  & -0.19   & $-$   & 41.10   \\
			HRank \cite{DBLP:conf/cvpr/LinJWZZ0020A25}       & 93.26  & 93.17  &  0.09  & 50.00  & 42.40   \\
			CNN-FCF \cite{li2019compressingA54}     & 93.14  & 93.38  & -0.24  & 43.09  & 42.78   \\
			NISP \cite{DBLP:conf/cvpr/Yu00LMHGLD18A29}        & $-$    & $-$    &  0.03   & 42.60  & 43.61   \\			
			Y.He et al. \cite{he2020learningA59}  & 93.59  & 93.72  & -0.13    & $-$   & 47.10   \\
			He et al. \cite{he2017channelA60}   & 92.80  & 91.80  &  1.00  & $-$   & 50.00   \\			
			AMC \cite{DBLP:conf/eccv/HeLLWLH18A34}         & 92.80  & 91.90  &  0.90  & $-$ & 50.00   \\
			DCP \cite{DBLP:conf/nips/ZhuangTZLGWHZ18A51}         & 93.80  & 93.59  &  0.21  & $-$   & 50.00   \\			
			DMC \cite{gao2020discreteA61}         & 93.62  & 93.69  & -0.07  & $-$   & 50.00   \\			
			SFP \cite{DBLP:conf/ijcai/HeKDFY18A23}         & 93.59  & 93.35  &  0.24   & $-$    & 52.60   \\		
			FPGM \cite{he2019filterA52}        & 93.59  & 92.89  &  0.70   & $-$ & 52.60   \\			
			CCP \cite{peng2019collaborativeA55}         & 93.50  & 93.42  &  0.08  & $-$  & 52.60   \\			
			Y.He et al. \cite{he2020learningA59}  & 93.59  & 93.34  &  0.25  & $-$  & 52.90   \\			
			SRR-GR \cite{wang2021convolutionalA58}       & 93.38  & 93.75  &  -0.37  & $-$  & 53.80   \\
			ABCPruner \cite{DBLP:conf/ijcai/LinJZZW020A36}   & 93.26  & 93.23  &  0.03   & 54.20  & 54.13   \\			
			EDP \cite{9246734A53}         & 93.61  & 93.61  &  0      & 54.18     & 57.71   \\
			ManiDP \cite{tang2021manifoldA27}             & 93.70  & 93.64  &  0.06      & $-$     & 62.40   \\
			\textbf{AIP($\textit{k}$=0.5)}  & \textbf{93.18}  & \textbf{93.51}  & \textbf{-0.33} & \textbf{56.47} & \textbf{65.15}   \\		
			\textbf{AIP($\textit{k}$=0.6)}  & \textbf{93.18}  & \textbf{92.79}  & \textbf{0.39} & \textbf{75.29} & \textbf{78.60}   \\		
			\hline 			
	\end{tabular}}	
	\label{tab2}	
\end{table*}

\subsection{Datasets}

CIFAR-10 contains 50,000 training images and 10,000 test images spanning 10 categories of objects. Each category contains 5,000 training images and 1,000 test images. On the CIFAR-10 dataset, we experiment with VGG-16, ResNet-56/110 and GoogLeNet.

CIFAR-100 contains 60,000 images spanning 100 categories of objects, and each category contains 600 pictures, of which 500 training images and 100 test images. On CIFAR-100, we evaluate our method with VGG-19 and ResNet-56.

ILSVRC-2012 is a large-scale dataset, which contains 1.28 million training images and 50000 validation images from 1000 classes. On ILSVRC-2012, we perform the experiments with ResNet-18.

PASCAL VOC is a benchmark dataset that contains 20 categories for object detection. VOC2007 contains 9,963 annotated images, including 5011 training images and 4952 test images, while VOC2012 contains 11,540 annotated images for object detection. On PASCAL VOC, we use VGG-16 as a backbone network to deploy SSD.

\begin{table*}[t]	
	\centering
	\caption{Performance comparison of ResNet-110 on CIFAR-10.}
	\renewcommand{\arraystretch}{1.1}
	{
		\begin{tabular}{c c c c c c c c}			
			\hline 
			Method  &Baseline Acc/\% & Pruned Acc/\% & Acc.drop/\% & Parameters/M  &  Parameters.drop/\%  &   FLOPs/M  &  FLOPs.drop/\% \\ 
			\hline
			\textbf{Baseline}         & 93.32  & $-$    & $-$    & 1.73        & $-$                & 256.04  & $-$           \\ 
			SFP \cite{DBLP:conf/ijcai/HeKDFY18A23}          & 93.68  & 93.83  & -0.15  & $-$   & $-$  & 216.00  & 14.60   \\
			Li et al. \cite{DBLP:conf/iclr/0022KDSG17A22}   & 93.53  & 93.55  & -0.02 & 1.68 & 2.30  & 213.00 & 15.90   \\
			Liu et al. \cite{DBLP:conf/iclr/LiuSZHD19A50}  & 93.14  & 93.22  & -0.08  & 1.68 & 2.30  & 213.00  & 15.90   \\
			\textbf{AIP($\textit{k}$=0.3)}  & \textbf{93.32}  & \textbf{94.43}  & \textbf{-1.11} & \textbf{1.46} & \textbf{27.00} & \textbf{195.32}  & \textbf{23.72}  \\
			SFP \cite{DBLP:conf/ijcai/HeKDFY18A23}         & 93.68  & 93.93  & -0.25   & $-$  & $-$ & 182.00  & 28.20   \\
			Li et al. \cite{DBLP:conf/iclr/0022KDSG17A22}   & 93.53  & 93.30  &  0.20  & 1.16  & 32.40  & 155.00  & 38.60   \\
			Liu et al. \cite{DBLP:conf/iclr/LiuSZHD19A50}  & 93.14  & 93.60  & -0.46   & 1.16  & 32.40  & 155.00  & 38.60   \\
			HRank \cite{DBLP:conf/cvpr/LinJWZZ0020A25}       & 93.50  & 94.23  & -0.73 & $-$  & 41.20   & $-$  & 39.40   \\
			SFP \cite{DBLP:conf/ijcai/HeKDFY18A23}         & 93.68  & 93.86  & -0.18   & $-$ & $-$  & 150.00   & 40.80   \\
			CNN-FCF \cite{li2019compressingA54}     & 93.58  & 93.67  & -0.09 & $-$  & 43.19  & $-$  & 43.08   \\
			NISP \cite{DBLP:conf/cvpr/Yu00LMHGLD18A29}        & $-$    & $-$    &  0.18  & $-$  & 43.25  & $-$  & 43.78   \\
			GAL \cite{DBLP:conf/cvpr/LinJYZCYHD19A14}         & 93.50  & 92.74  &  0.76 & 0.95   & 44.80 & 130.20  & 48.50   \\
			FPGM \cite{he2019filterA52}         & 93.68  & 93.73  & -0.05  & $-$ & $-$  & 121.00 & 52.30   \\
			Y.He et al. \cite{he2020learningA59} & 93.68  & 93.79  & -0.11  & $-$ & $-$  & 101.00   & 60.30   \\
			\textbf{AIP($\textit{k}$=0.5)}  & \textbf{93.32}  & \textbf{94.12}  & \textbf{-0.80} & \textbf{0.71} & \textbf{58.96} & \textbf{97.29}  & \textbf{62.00}  \\	
			ABCPruner \cite{DBLP:conf/ijcai/LinJZZW020A36}   & 93.50  & 93.58  & -0.08 & 0.56   & 67.41 & 89.87   & 65.04   \\
			HRank \cite{DBLP:conf/cvpr/LinJWZZ0020A25}       & 93.50  & 92.65  &  0.85   & $-$   & 68.60   & $-$   & 68.70   \\
			CNN-FCF \cite{li2019compressingA54}     & 93.58  & 92.96  &  0.62   & $-$   & 69.51   & $-$   & 70.81   \\			
			\textbf{AIP($\textit{k}$=0.7)}  & \textbf{93.32}  & \textbf{93.98}  & \textbf{-0.66} & \textbf{0.45} & \textbf{73.99} & \textbf{62.27}  & \textbf{75.32}  \\			
			\hline 			
	\end{tabular}}	
	\label{tab3}
\end{table*}

\begin{table*}[t]	
	\centering
	\caption{Performance comparison of GoogLeNet on CIFAR-10.}
	\renewcommand{\arraystretch}{1.1}
    {
		\begin{tabular}{c c c c c c c c}			
			\hline 
			Method  &Baseline Acc/\% & Pruned Acc/\% & Acc.drop/\% & Parameters/M  &  Parameters.drop/\%  &   FLOPs/G  &  FLOPs.drop/\% \\ 
			\hline
			\textbf{Baseline}         & 94.72  & $-$    & $-$    & 6.17        & $-$                & 1.53  & $-$           \\ 
			\textbf{AIP($\textit{k}$=0.4)}  & \textbf{94.72}  & \textbf{95.25}  & \textbf{-0.52} & \textbf{4.08} & \textbf{33.87} & \textbf{0.95}  & \textbf{37.95}  \\
			\textbf{AIP($\textit{k}$=0.5)}  & \textbf{94.72}  & \textbf{95.13}  & \textbf{-0.41} & \textbf{3.19} & \textbf{48.30} & \textbf{0.74}  & \textbf{52.04}  \\
			ABCPruner \cite{DBLP:conf/ijcai/LinJZZW020A36}                   & 95.05  & 94.84  & 0.21 & 2.46 & 60.14  & 0.51  & 66.56   \\
			\textbf{AIP($\textit{k}$=0.7)}  & \textbf{94.72}  & \textbf{95.06}  & \textbf{-0.34} & \textbf{2.12} & \textbf{65.64} & \textbf{0.47}  & \textbf{69.34}  \\			
			\hline 			
	\end{tabular}}	
	\label{tab4}	
\end{table*}

\subsection{Results Comparison on CIFAR-10}

We first prune VGG-16 on the CIFAR-10, and the results are shown in TABLE \ref{tab1}. It can be seen from the table that when $k$=0.3, our method reduces up to 47.86\% of the parameters and 44.39\% of the FLOPs for VGG-16, however, the accuracy of the network is even improved by 0.54\% compared with the baseline. When the network compression ratio exceeds 80\%, the compact network still has a performance improvement of 0.17\%. Although the parameter compression ratio of ABCPruner \cite{DBLP:conf/ijcai/LinJZZW020A36} is 5.65\% higher than that of our method, the pruning rate of FLOPs is lower than that of this paper and the final accuracy after pruning is also smaller (-0.06\% vs. -0.17\%). As the VGG-16 continues to be compressed, the accuracy of the network is gradually declining. When discarding 97.22\% of the parameters and 96.57\% of the FLOPs with $k$=0.7, the accuracy of the final network still reaches 90.29\%.

The experimental results show that for the CIFAR-10, the VGG-16 does have a certain degree of parameter redundancy. Compressing the network can reduce the impact of overfitting and improve the accuracy of the network. At the same time, the effectiveness of the pruning method proposed in this paper is preliminarily verified. Then, we continue to cut ResNet-56, and the experimental results are tabulated in TABLE \ref{tab2}. When $k$=0.4, the parameters and FLOPs of ResNet-56 are reduced by 37.65\% and 46.36\%, respectively. At this time, the accuracy of the network after pruning is increased by 0.82\%. And when $k$=0.5, the pruning rate has exceeded 50.00\%, but the network still has a performance improvement of 0.33\%, which is significantly better than the compared algorithms. Although the final accuracy improvement of SRR-GR \cite{wang2021convolutionalA58} is 0.04\% higher than that of ours, the compression rate of its FLOPs is relatively low by 11.35\% (53.80\% vs. 65.15\%). The accuracy of ResNet-56 only drops by 0.39\% when deleting 75.29\% of the parameters and 78.60\% of the FLOPs. In this case, the network parameters are only 0.21M. In addition, it can be found that when $k$=0.4, the classification accuracy of ResNet-56 is 94.00\%, which is 0.23\% higher than that of VGG-16 when $k$=0.7, however, the parameters are only about 1/5 of VGG-16. It also confirms from the side that the residual module can effectively improve the performance of CNNs in image classification tasks.

Then, we compress ResNet-110. From TABLE \ref{tab3}, it can be concluded that the baseline accuracy of ResNet-110 on the CIFAR-10 is 93.32\%. When 27.00\% of the parameters and 23.72\% of FLOPs are discarded, the accuracy increased by 1.11\%. HRank \cite{DBLP:conf/cvpr/LinJWZZ0020A25} compresses parameters and FLOPs by 41.20\% and 39.40\%, respectively, which is about 20\% lower than our method when $k$=0.5, and the performance is also 0.07\% worse. The performance of the compressed network is still improved by 0.66\% compared to the original network even when 73.99\% of the parameters and 75.32\% of the FLOPs are eliminated. At this time, the network scale is similar to that of CNN-FCF \cite{li2019compressingA54}, but the final accuracy loss is 1.28\% lower (-0.66 vs. 0.62). This experiment shows that ResNet-110 has obvious parameter redundancy on the CIFAR-10, which leads to overfitting during the training process, resulting in lower accuracy of the original network. And our method can achieve better accuracy recovery in the case of accurately compressing the ResNet-110. It also manifests that the compressing rate of parameters and FLOPs and the accuracy drop using our pruning method are significantly better than all comparative methods.

In order to further demonstrate the applicability of AIP to various convolutional networks, we continue to prune GoogLeNet. The experimental results are depicted in TABLE \ref{tab4}. Due to the Inception module, GoogLeNet increases the width, therefore the baseline accuracy on the CIFAR-10 reaches 94.72\%, which is ahead of VGGNet and ResNet. When $k$=0.4, after deleting 33.87\% of the parameters and 37.95\% of the FLOPs, the accuracy of the network increased by 0.52\%. When the parameters and FLOPs are compressed to about 50\%, the performance of the compact network is increased by 0.41\%. Even if the parameters and the FLOPs are removed by 65.64\% and 69.34\% respectively, the classification accuracy of the network is still improved by 0.34\%. It attests that GoogLeNet is also redundant on the CIFAR-10. Using the iterative pruning method in this paper can effectively eliminate unimportant parameters and improve the experimental performance of GoogLeNet.

\begin{table*}[t]	
	\centering
	\caption{Performance comparison of VGG-19 on CIFAR-100.}
	\renewcommand{\arraystretch}{1.1}
	{
		\begin{tabular}{c c c c c c c c}			
			\hline 
			Method  &Baseline Acc/\% & Pruned Acc/\% & Acc.drop/\% & Parameters/M  &  Parameters.drop/\%  &   FLOPs/M  &  FLOPs.drop/\% \\ 
			\hline
			\textbf{Baseline}         & 72.01  & $-$    & $-$    & 20.09        & $-$                & 399.52  & $-$           \\ 
			\textbf{AIP($\textit{k}$=0.3)}  & \textbf{72.01}  & \textbf{74.42}  & \textbf{-2.41} & \textbf{9.39} & \textbf{53.26} & \textbf{215.07}  & \textbf{46.17}  \\
			\textbf{AIP($\textit{k}$=0.4)}  & \textbf{72.01}  & \textbf{73.79}  & \textbf{-1.78} & \textbf{5.49} & \textbf{72.67} & \textbf{134.98}  & \textbf{66.21}  \\
			Slimming \cite{ISI:000425498402086A24}                   & 73.26  & 73.48  & -0.22 & 5.00 & 75.10  & 251.00  & 37.10   \\
			Liu et al. \cite{DBLP:conf/iclr/LiuSZHD19A50}  & 72.63  & 73.08  & -0.45 & 5.00 & 75.10  & 251.00  & 37.10   \\
			\textbf{AIP($\textit{k}$=0.5)}  & \textbf{72.01}  & \textbf{72.58}  & \textbf{-0.57} & \textbf{3.00} & \textbf{85.07} & \textbf{74.17}  & \textbf{81.44}  \\			
			\hline 			
	\end{tabular}}	
	\label{tab5}	
\end{table*}

\begin{table*}[t]	
	\centering
	\caption{Performance comparison of ResNet-56 on CIFAR-100.}
	\renewcommand{\arraystretch}{1.1}
	{
		\begin{tabular}{c c c c c c}			
			\hline 
			Method  &Baseline Acc/\% & Pruned Acc/\% & Acc.drop/\%  & FLOPs/M   & FLOPs.drop/\% \\ 
			\hline
			Baseline        & 71.36  & $-$  & $-$    & 127.09    & $-$   \\
			\textbf{AIP($\textit{k}$=0.3)}  & \textbf{71.36}  & \textbf{73.57}  & \textbf{-2.21}  & \textbf{92.35} & \textbf{27.33}   \\
			\textbf{AIP($\textit{k}$=0.4)}  & \textbf{71.36}  & \textbf{71.88}  & \textbf{-0.52}  & \textbf{65.18} & \textbf{48.71}   \\		
			Y.He et al. \cite{he2020learningA59}  & 71.41  & 70.83  & 0.58    & 60.80   & 51.60   \\			
			SFP \cite{DBLP:conf/ijcai/HeKDFY18A23}         & 71.40  & 68.70  &  2.61  & 59.40   & 52.60   \\		
			FPGM \cite{he2019filterA52}        & 71.41  & 69.66  &  1.75   & 59.40 & 52.60   \\			
			\textbf{AIP($\textit{k}$=0.5)}  & \textbf{71.36}  & \textbf{71.18}  & \textbf{0.18} & \textbf{40.33} & \textbf{68.27}   \\		
			\hline 				
	\end{tabular}}	
	\label{tab6}	
\end{table*}

\begin{table*}[t]	
	\centering
	\caption{Performance comparison of ResNet-18 on ILSVRC-2012.}
	\renewcommand{\arraystretch}{1.1}
	{
		\begin{tabular}{c c c c c c c}			
			\hline 
			Method  & Top-1 Acc/\%   & Top-1 Acc.drop/\%   & Top-5 Acc/\%   & Top-5 Acc.drop/\% & Parameters.drop/\% & FLOPs.drop/\% \\ 
			\hline
			MIL \cite{dong2017moreA63}    & 66.33   & 3.43   & 86.94   & 2.14    & $-$   & 33.30    \\
			DSA \cite{ning2020dsaA65}    & 68.61   & 1.11   & 88.35   & 0.72    & $-$   & 40.00    \\
			SFP \cite{DBLP:conf/ijcai/HeKDFY18A23}   & 67.10   & 3.18   & 87.78   & 1.85    & $-$   & 41.80    \\
			FPGM \cite{he2019filterA52}    & 68.41   & 1.35   & 88.48   & 0.60    & $-$   & 41.80    \\
			PFP \cite{DBLP:conf/iclr/LiebenweinBLFR20A64}    & 65.65   & 4.11   & 86.75   & 2.33    & $-$   & 43.00    \\
			\textbf{AIP($\textit{k}$=0.5)}  & \textbf{69.36} & \textbf{0.66}  & \textbf{88.71}  & \textbf{0.52}  & \textbf{36.78}  & \textbf{45.55}   \\
			ABCPruner \cite{DBLP:conf/ijcai/LinJZZW020A36}    & 67.28   & 2.38   & 87.67   & 1.41    & 43.55   & 44.88    \\
			FBS \cite{DBLP:conf/iclr/GaoZDMX19A62}    & 68.17   & 1.59   & 88.22   & 0.86    & $-$   & 49.50    \\
			ManiDP \cite{tang2021manifoldA27}    & 68.88   & 0.88   & 88.76   & 0.32    & $-$   & 51.00    \\
			\textbf{AIP($\textit{k}$=0.7)}  & \textbf{67.35}  & \textbf{2.67}  & \textbf{87.79}  & \textbf{1.44}   & \textbf{58.51}  & \textbf{65.07}   \\			
			\hline 			
		\end{tabular}
	}	
	\label{tab7}	
\end{table*}

\subsection{Results Comparison on CIFAR-100}

We continue to prune VGG-19 and ResNet-56 on the CIFAR-100, and the experimental results are reported in TABLE \ref{tab5} and TABLE \ref{tab6}, respectively. CIFAR-100 has the same total number of training and test images as CIFAR-10, but the category has increased from 10 to 100. As the training data for each class of images decreases, the performance of the convolutional neural network also drops significantly. It can be seen from TABLE \ref{tab5} that the baseline accuracy of VGG-19 on CIFAR-100 is only 72.01\%. When pruning 72.67\% of the parameters and 66.21\% of the FLOPs, the accuracy of the retrained compact network is increased by 1.78\%. Even when $k$=0.5, the performance is still improved by 0.57\% when parameters and FLOPs are compressed by 85.07\% and 81.44\%, respectively. And it is significantly better than Slimming \cite{ISI:000425498402086A24} and Liu et al. \cite{DBLP:conf/iclr/LiuSZHD19A50} in terms of network compression ratio and performance recovery. This manifests that our AIP is also applicable to datasets with relatively few training samples. TABLE \ref{tab6} shows that the baseline accuracy of ResNet-56 is 71.36\%. Because parameters and FLOPs of ResNet-56 are significantly less than VGG-19, the redundancy of ResNet-56 is also smaller. However, when the number of FLOPs is discarded by 48.71\%, there is still a 0.52\% improvement in performance. When $k$=0.5, we remove 68.27\% of the FLOPs with 0.18\% accuracy drop that is still significantly superior to the comparison method.

Experiments on the CIFAR datasets preliminarily verify the effectiveness and superior performance of the proposed method in image classification tasks. Our AIP can achieve a certain degree of compression for the parameters and FLOPs of VGGNet, ResNet, and GoogLeNet almost without accuracy drop. It also fully indicates that in different tasks, the existing CNNs have certain parameter redundancy, and removing these unimportant parameters can achieve network compression and acceleration without affecting the performance of networks. In this way, the computational cost of the neural network will reduce remarkably.

\begin{table*}[htbp]	
	\centering
	\caption{The results of pruning SSD on PASCAL VOC.}
	\renewcommand{\arraystretch}{1.1}
	{
		\begin{tabular}{c c c c c c c}			
			\hline 
			Method  & mAP/\%   & mAP.drop/\%    & Parameters/M  &  Parameters.drop/\%  &   FLOPs/G  &  FLOPs.drop/\%  \\ 
			\hline
			
			Baseline  & 76.10   & $-$   & 26.29   & $-$    & 11.34   & $-$ \\
			
			\textbf{AIP($\textit{k}$=0.3)}  & \textbf{75.60} & \textbf{0.50}  & \textbf{13.76}  & \textbf{47.66}  & \textbf{7.91}  & \textbf{30.25}   \\
			
			\textbf{AIP($\textit{k}$=0.4)}  & \textbf{75.20} & \textbf{0.90}  & \textbf{11.13}  & \textbf{57.66}  & \textbf{5.21}  & \textbf{54.06}   \\			
			\hline 			
	\end{tabular} }	
	\label{tab8}	
\end{table*}

\subsection{Results Comparison on ILSVRC-2012}

To further assess the effectiveness of the proposed pruning method, we experiment on the large image classification dataset ILSVRC-2012 with 1000 categories which are difficult to precisely classify, and the parameters of the CNNs are less redundant, so pruning is more challenging. In this subsection, we select ResNet-18 with fewer parameters and FLOPs for pruning, which can highlight the power of our method. As we can see from TABLE \ref{tab7} that when pruning less than 45.00\% of the FLOPs via AIP, the Top-1 and Top-5 accuracy loss is smaller than that of other methods. Although the parameter compressing rate of ABCPruner \cite{DBLP:conf/ijcai/LinJZZW020A36} is 6.77\% higher than that of ours, and the FLOPs pruning rate is 0.67\% lower, the performance drop after pruning is significantly greater. The Top-1 accuracy in \cite{DBLP:conf/ijcai/LinJZZW020A36} loses 2.38\%, while the performance only drops 0.66\% via our method, and its Top-5 accuracy also decreases 0.89\% higher than AIP. FBS \cite{DBLP:conf/iclr/GaoZDMX19A62} pruning 3.95\% FLOPs higher than that of AIP, and its Top-1 and Top-5 accuracy loss is also higher than ours by 0.93\% (1.59\% vs. 0.66\%) and 0.34\% (0.86\% and 0.52\%) respectively. When $k$=0.5, the cutting rate of FLOPs using AIP is 5.45\% lower than that of ManiDP \cite{tang2021manifoldA27}, and the Top-5 accuracy drop is 0.20\% higher (0.52\% vs. 0.32\%), but the Top-1 accuracy loss is 0.22\% lower (0.66\% vs. 0.88\%). When $k$=0.7, AIP deletes 58.51\% of the parameters and 65.07\% of the FLOPs. In this scenario, the compression degree is significantly higher than the comparative pruning algorithms, and the accuracy loss is also higher. To the best of our knowledge, this is because the number of remaining parameters is too little to adequately extract the target information in the images during the learning process with the continuous compression of the network, and it results in the decrease of the final classification performance.

All the experiments for image classification reveal that our iterative pruning method can achieve a similar degree of compression rate on the parameters and FLOPs of convolutional networks. For simple tasks, after using AIP for network pruning, overfitting is eliminated, and the performance of the compact network can maintain or even exceed the accuracy of the original network after retraining. The complex classification task requires more parameters to extract the semantic information in the image. There are almost no redundant parameters in tiny convolutional networks, therefore pruning will be accompanied by a decrease in accuracy. However, our AIP can still control the performance loss in a smaller range. It indicates that the adversarial iterative pruning method based on knowledge transfer proposed in this paper can effectively remove unimportant parameters in the CNNs and reduce network redundancy in the sense that it also has regularization on the network training.

\begin{figure*}[t]
	
	\centering
	
	\includegraphics[width=18cm]{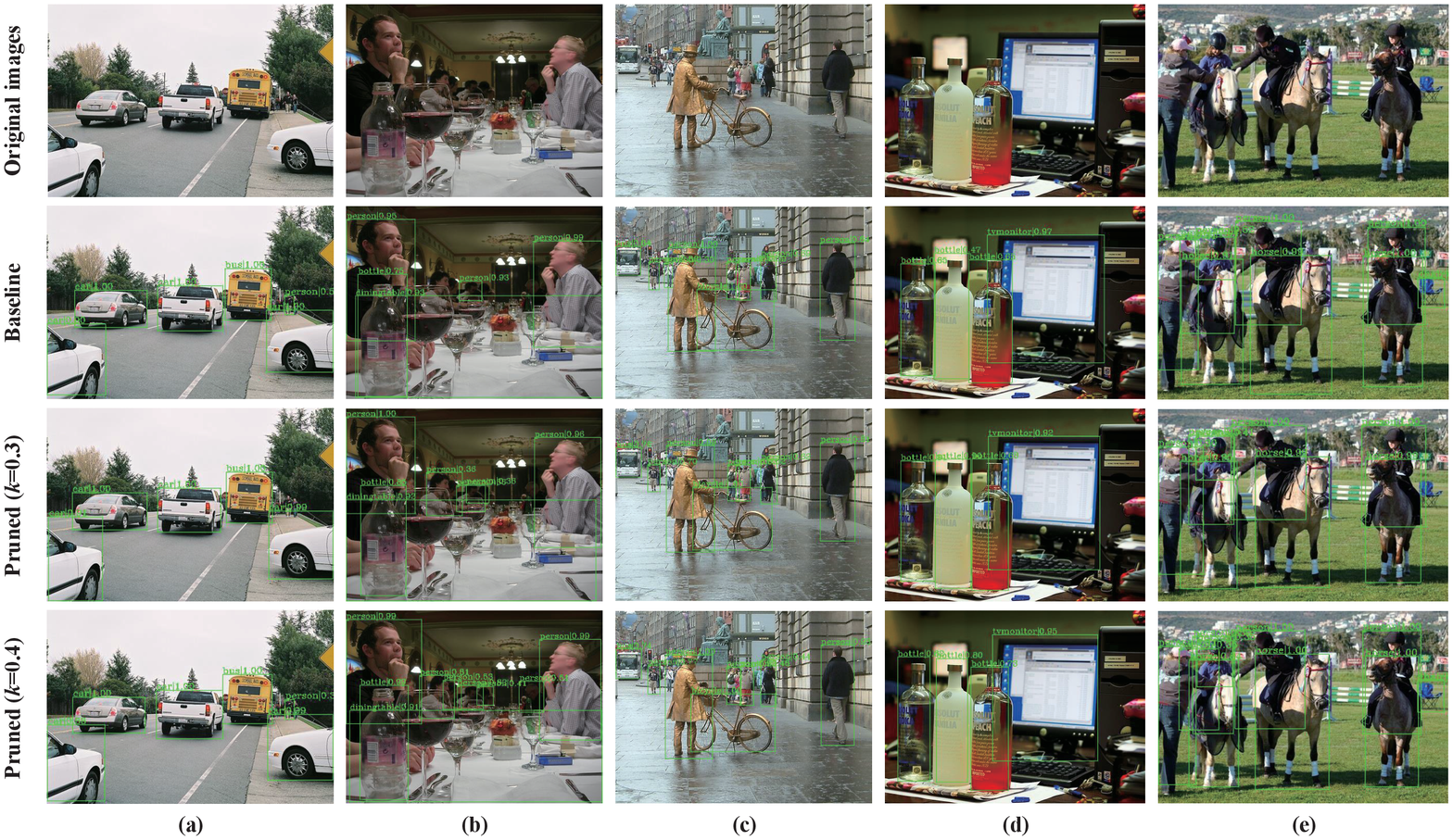} 
	
	\caption{Visualization of pruning SSD on the PASCAL VOC.}
	\label{img5}
	
\end{figure*}

\begin{figure}[t]
	
	\centering
	
	\includegraphics[width=8cm]{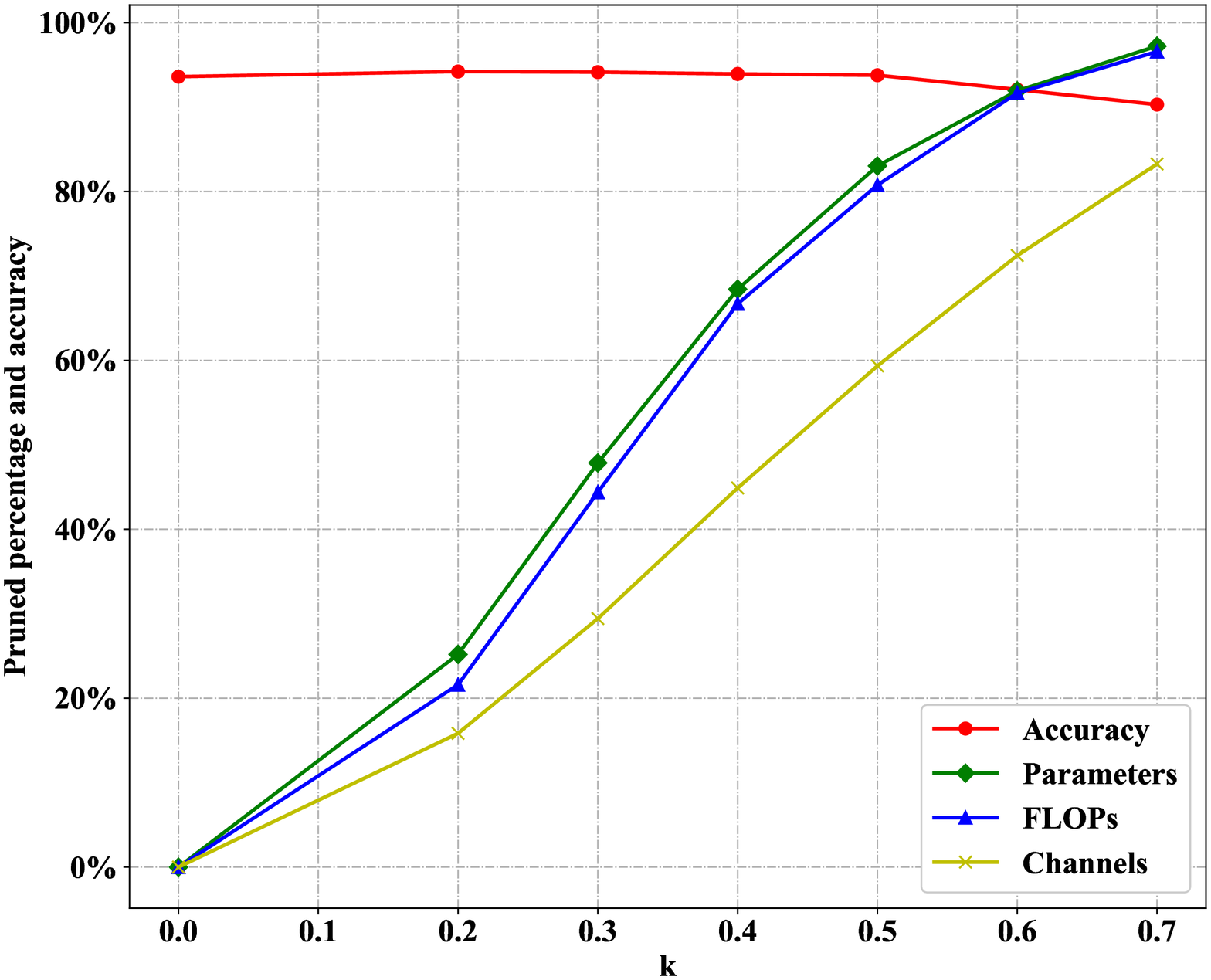} 
	
	\caption{The influence of k on the accuracy and pruning rate of VGG-16 on the CIFAR-10.}
	\label{img6}
	
\end{figure}

\subsection{Pruning SSD on PASCAL VOC}

The existing network pruning algorithms are almost totally for single-target image classification tasks with obvious targets and rarely involve other more complex tasks. In the real world, the scenarios of object detection are more extensive and the requirements for low storage and real-time are higher. However, in the context of uncertain conditions such as occlusion, size, and light changes, these tasks often need more complicated models. Therefore, compressing the model for object detection while maintaining accuracy faces salient challenges. To show off the generalization of the proposed method, we prune the SSD on the PASCAL VOC object detection dataset. The backbone of the SSD adopts the VGG-16 trained on the CIFAR-100. Here, we compare the parameters and FLOPs compressing rate and Mean Average Precision (mAP) loss. The results are depicted in TABLE \ref{tab8}. When $k$=0.3, the pruning rates of the parameters and FLOPs are 47.66\% and 30.25\%, respectively. Compared with the mAP of the baseline of 76.10\%, the detection accuracy of the compact SSD only decreases by 0.50\%. While pruning 57.66\% of the parameters and 54.06\% of the FLOPs in the SSD with $k$=0.4, the mAP drops by 0.90\%.

To visually display the results of the pruned SSD in the object detection, we select five pictures in PASCAL VOC to visualize the experiments in TABLE \ref{tab8}. And the results are depicted in Fig.\ref{img5}. The first line is the original images, and the second line is the detection result obtained using the baseline SSD, while the last two lines are the pruned results via AIP with $k$=0.3 and $k$=0.4. It can be found from the figure that the compressed SSD can still correctly detect the object in the images, despite the position and size of the detection frame may alter slightly within an acceptable range. Moreover, the confidence of some targets will also fluctuate to a certain extent. For example, the baseline confidence of the tvmonitor in figure (d) is 0.97, but when $k$=0.3 and $k$=0.4, they are 0.92 and 0.95, respectively. The confidence of some targets in the other pictures is also different. We conjecture this is due to the detection accuracy of some categories has been improved after pruning, although the overall mAP is slightly lower. To be more specific, the detection precision of some targets will even improve when compressing the network. For instance, the baseline confidence of the bottle in figure (b) is 0.75, however, when $k$=0.3 it reaches 0.85, and when $k$=0.4 it even increases to 0.99. It also reveals that pruning can improve the capability of the network recognition for some target classes by reducing model redundancy. Moreover, when $k$=0.4, the SSD is compressed by more than 50\%. At this time, even more persons are accurately found than the baseline in figure(b). It manifests that the ability to distinguish people has been developed. The above experiments further verify that the pruning method in this paper also has good generalization in the field of object detection.

\subsection{Ablation Analysis}

Then, we conduct the ablation analysis on the proposed AIP method. This section is composed of the following three parts: the influence of $k$ on the pruning ratio, the influence of $k$ on the compression magnitude in different layers, and the influence of three modules of attention map transfer, knowledge distillation, and adversarial training on the network performance recovery.

\subsubsection{The Influence of $k$ on The Pruning Rate}

The pruning threshold factor $k$ is the parameter used to adjust the compression ratio in our proposed pruning algorithm. The larger the $k$, the greater the pruning threshold, so that the higher the degree of network compression. To reveal the influence of the $k$, we perform six groups of pruning experiments on VGG-16 by setting different $k$ on the CIFAR-10, and the results are shown in Fig.\ref{img6}. It can be seen from the figure that as the $k$ increases, the pruning rate of parameters, FLOPs, and channels constantly exceeds. When the value of $k$ is small, the cropping ratio rises faster meanwhile the curve is relatively steep. But with the continuous growth of $k$, the curve of the parameters and the FLOPs gradually tends to be smooth, while the compressing rate of channels almost still linearly rises. From the figure, it is clear that when the compression ratio of parameters and FLOPs are less than 80\%, and that of the number of channels is less than 60\%, the accuracy of the pruned network remains unchanged or even slightly improved compared to the baseline. Nevertheless, the performance of the compact network begins to decline if continues to compress. This is because when pruning fewer parameters, the redundancy and the impact of overfitting are reduced so that the performance will be improved. But when removing too many parameters, the network is difficult to cope with the classification tasks which causes performance degradation.

\begin{figure*}[t]
	
	\centering
	
	\includegraphics[width=18cm]{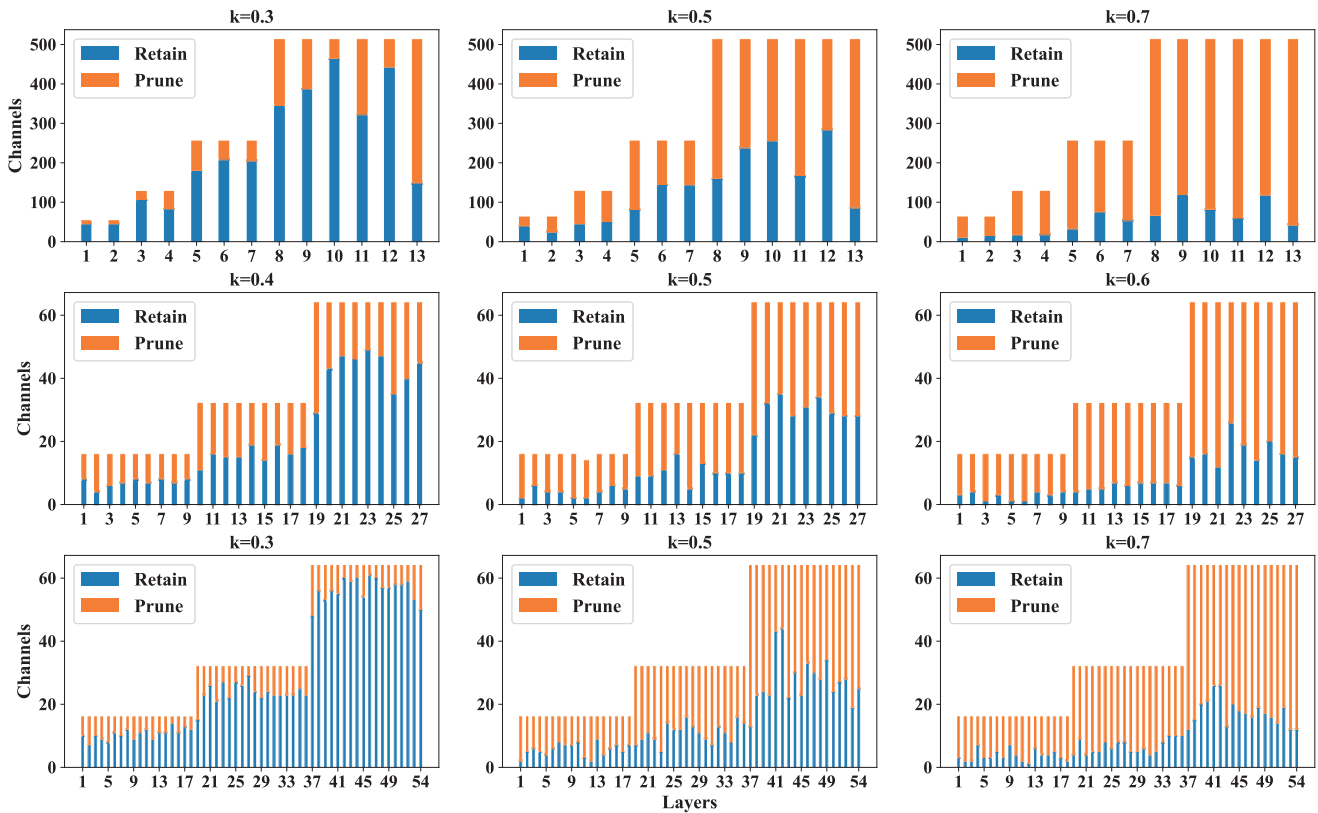} 
	
	\caption{The influence of k for channels of VGG-16 (Top), ResNet-56 (Second row), and ResNet-110 (Bottom) on CIFAR-10.}
	\label{img7}
	
\end{figure*}

\subsubsection{The Influence of $k$ on The Compression Magnitude in Different layers}

To better show the compression amplitude of each layer of the network under different pruning ratios, this section visualizes the number of channels of VGG-16, ResNet-56, and ResNet-110 in CIFAR-10 as Fig.\ref{img7}. The three rows from top to bottom are VGG-16, ResNet-56, and ResNet-110. It can be found from the first row that the last layer of VGG-16 has the most remarkable redundancy. The last layer has eliminated more than 50.00\% of the channels with $k$=0.3, while the discarding ratio of the network is small. As the pruning ratio increases, the number of retained channels in the 9-th and 12-th layers is more than that in the other layers. It implies that the impact of the two layers on extracting target information is more pivotal than that of other layers. For ResNet-56, the number of channels reserved in the 23-th layer is more than that of the 22-th layer with little compression. But when $k$=0.6, the cropping ratio raises, the number of channels saved in the 22-th layer is significantly more than the other layers. It indicates that as the pruning rate changes, the importance of different layers also varies to improve the performance as much as possible. At the same time, it reiterates that the pruning strategy proposed in this paper can adaptively adjust the pruning range of each layer according to different compression rates to obtain a compact network that meets the performance requirements.

\subsubsection{The Influence of Three Modules on The Performance Recovery}

To analyze the influence of attention map transfer, knowledge distillation, and adversarial training on the performance recovery of the compact network, we retrain the pruned VGG-16 via different strategies on the CIFAR-10. The number of training epochs and other hyperparameter settings remain the same. The results are tabulated in TABLE \ref{tab9}. As we can see from the table that the network trained by all the three strategies has the highest accuracy, which can reach 94.14\%. The accuracy obtained using the attention map transfer and knowledge distillation is 94.10\%, which is only 0.04\% lower than applying the three strategies. When utilizing knowledge distillation and adversarial training, the accuracy is 94.06\%. However, the final accuracy is only 93.96\% which is even 0.07\% lower than the performance using knowledge distillation alone when adopting attention map transfer and adversarial training. Therefore, knowledge distillation plays the most considerable role in the three modules, and knowledge distillation combines attention map transfer can achieve superior performance as applying the three modules. Accordingly, we only make use of attention map transfer and knowledge distillation in our retraining phase. It can also be seen from the table that the performance via the adversarial training alone is better than only using the attention map transfer. It is because the attention map only works on the intermediate output feature maps and does not restrict the final output, while the adversarial game can directly optimize the output features, so it can better improve the performance of the pruned network.

\begin{table}[t]	
\centering
\caption{Performance comparison of retraining pruned VGG-16 on CIFAR-10.}
\renewcommand{\arraystretch}{1.1}
{
	\begin{tabular}{c c c c c c}			
		\hline 
		Method  & ${\mathcal{L}}_{AT}$   & ${\mathcal{L}}_{KD}$    & ${\mathcal{L}}_{D}$  &   Acc/\%  &  Acc.drop/\%  \\ 
		\hline
		Baseline  & $-$   & $-$   & $-$   & 93.60    & $-$ \\
		\hline
		\multirow{7} {*} {AIP($\textit{k}$=0.3)} 
		
		& \checkmark &     &     & 93.68   & -0.08 \\
		
		&     & \checkmark &     & 94.03   & -0.43 \\
		
		&     &     & \checkmark & 93.88   & -0.28 \\
		
		& \checkmark &     & \checkmark   & 93.96   & -0.36 \\
		
		&   & \checkmark   & \checkmark   & 94.06   & -0.46 \\
		
		& \checkmark & \checkmark   &     & 94.10   & -0.50 \\
		
		& \checkmark & \checkmark   & \checkmark   & \textbf{94.14}   & \textbf{-0.54}	\\ 	
		\hline 			
	\end{tabular} }
	
\label{tab9}	
\end{table}

\section{Conclusion and Future Work}

In this paper, we propose an adversarial iterative network pruning method based on knowledge transfer. The attention maps generated by the unpruned network are used to guide the pruned network to pay more attention to the classification target. And then, the output features of the compressed network are constrained and corrected by knowledge distillation. Meanwhile, we introduce the idea of the adversarial game, so that the original network and the pruned network can compete through a discriminator formed by a shallow neural network. In this way, the performance of the compressed network can be recovered as soon as possible after each pruning step. Together with the above optimizing methods, we design an iterative channel pruning method based on the importance of feature maps. The final compact network obtained will restore the accuracy through retraining from scratch. We conduct extensive experiments for pruning VGGNet, ResNet, and GoogLeNet on image classification datasets of CIFAR-10, CIFAR-100, and ILSVRC-2012. The results have manifested that AIP is comparable with state-of-the-art network pruning methods in performance and pruning rate of parameters and FLOPs. On CIFAR-10, after compressing 75.29\% of the parameters and 78.60\% of the FLOPs of ResNet-56, the accuracy only drops by 0.39\%. In the object detection task PASCAL VOC, when removing more than 50\% of the parameters and FLOPs of the SSD, the mAP is only reduced by 0.9\%. The final ablation analysis reveals that the pruning factor can achieve flexible control of the compression rate. The experiments fully show that the proposed can be widely applied in different CNNs, image datasets, and various computer vision tasks.

In the future, we will integrate the channel pruning method with other compression schemes such as quantization. Furthermore, we will consider applying existing approaches to accelerate other real-world vision tasks and even natural language processing.

\bibliography{mybibfile}

\end{document}